\documentclass[10pt]{IEEEtran}
\IEEEoverridecommandlockouts

\usepackage{graphicx}
\usepackage{authblk}
\usepackage[english]{babel}
\usepackage{algorithm2e}
\usepackage{caption}
\usepackage{subcaption}
\usepackage{enumerate}
\usepackage{color}
\usepackage{amsmath,amssymb,amsfonts}
\DeclareMathOperator*{\argminB}{argmin}
\DeclareMathOperator*{\argmaxB}{argmax}

\begin{document}

\title{Segmented and Non-Segmented Stacked Denoising Autoencoder for Hyperspectral Band Reduction}

\author[1]{Muhammad Ahmad}
\author[2]{Asad Khan}
\author[1]{Adil Mehmood Khan}
\author[3]{Rasheed Hussain}

\affil[1]{Institute of Robotics, Innopolis University, Innopolis, 420500, Kazan, Tatarstan, Russia.}
\affil[2]{Graphic and Computing Lab, School of Computer Science, Normal University, Guangzhou, China.}
\affil[3]{Institute of Information Systems, Innopolis University, Innopolis, 420500, Kazan, Tatarstan, Russia}

\maketitle

\begin{abstract}
Hyperspectral image analysis often requires selecting the most informative bands instead of processing the whole data without losing the key information. Existing band reduction (BR) methods have the capability to reveal the nonlinear properties exhibited in the data but at the expense of loosing its original representation. To cope with the said issue, an unsupervised non-linear segmented and non-segmented stacked denoising autoencoder \textit{(UDAE)} based BR method is proposed. Our aim is to find an optimal mapping and construct a lower-dimensional space that has a similar structure to the original data with least reconstruction error. The proposed method first confronts the original hyperspectral data into smaller regions in spatial domain and then each region is processed by \textit{UDAE} individually. This results in reduced complexity and improved efficiency of BR for both semi-supervised and unsupervised tasks, i.e. classification and clustering. Our experiments on publicly available hyperspectral datasets with various types of classifiers demonstrate the effectiveness of \textit{UDAE} method which equates favorably with other state-of-the-art dimensionality reduction and BR methods.
\end{abstract}

\section{Introduction and Related Work}
Remote sensing deals with the extraction of meaningful information of objects of interests from the earth surface based on their radiance acquired through the remote sensors. One of the applications of remote seining is hyperspectral imaging. Hyperspectral imaging has been extensively used in many practical applications such as monitoring of land changes, urban development, environmental and mineral exploration [1]. 

Due to the enhanced capabilities of hyperspectral sensors, these sensors have the ability to capture hundreds of narrow and contiguous spectral bands with high resolutions. This high resolution lead to better identification of relatively small structures. Owing to the high resolutions, the geometrical structure of a scene has a great perceptual significance that can be directly exploited for modeling the objects; however, it usually leads to the curse of dimensionality (CoD) (CoD refers to various phenomena that arise when analyzing data in high-dimensional spaces, often with hundreds or thousands of dimensions) [2]. To cope with high dimensionality issue, several linear and non-linear transformation-based band reduction (BR) methods have been proposed [3].

These methods include principal component analysis (PCA) [4], kernel principal component analysis (KPCA) [5], linear discriminant analysis (LDA) [6], stepwise linear discriminant analysis (SWLDA) [7][8], kernel linear discriminant analysis (KLDA) [9], factor analysis (FA), minimum noise fraction transformation and genetic algorithms [10]. All these methods change the physical meaning of the original hyperspectral space because the channels in lower dimensional space do not correspond to the individual band but to their linear combinations. BR methods to select a subset of the original space may also be preferred when the physical meaning of bands needs to be maintained [11]. Furthermore, in order to use the existing BR methods, one must determine the number of bands, the metric or criterion for selection and the search strategy for optimum performance. Several methods have been proposed to select the optimum values of these parameters [13][14]. These methods include virtual dimensionality (VD) and particle swarm optimizer (PSO).

The concept of VD is used to estimate the number of spectrally distinct signatures that serve as a reference value [15] and PSO method to automatically determine the optimal number of bands [16][17]. Several BR methods take distance metrics as class separabilities. These metrics include Jeffrey’s Matusita [18], Bhattacharyya [19] and divergence-based distance metrics [20]. However, a representative spectral signature for each class can be used to compute the minimum spatial covariance [21]. Furthermore, sometimes the accuracy of a classifier can directly be used as objective function [22]. The core objective of any searching strategy such as forward searching methods [19] is to avoid testing all possible band combinations. A firefly method [23] is an evolutionary computational technique which has also been used for BR [20][24]. However, BR with firefly searching method can be expensive because classification needs to be done during the BR process. It is computationally expensive if the selected classifier is complex with many training samples. Moreover, the selected bands may be optimal only for the involved classifier.

On the other hand an autoencoder (AE) [12] can be used to learn a compact representation of the input data; one that also ensures a minimum difference between corresponding inter-point relations in two spaces. A transformation found by an AE is regarded as preferable if it minimizes the reconstruction error and ensures that the encoding and decoding does not affect the structure embedded in the data. In other words, the importance of preserving relations between points is emphasized. Therefore, unlike traditional unsupervised linear as well as non-linear BR methods, the transformation found by a deep autoencoder (DAE) may be both compact and reflect the structure present in the original hyperspectral data at the same time. Therefore, we address the above-mentioned issues using an unsupervised non-linear segmentation-based deep autoencoder \textit{(UDAE)} method. Our proposed UDAE is classifier independent and significantly improves the classification performance and decrease the computational complexity. Furthermore, we employ several classifiers to validate its performance. To this end, our aim is not only to find an optimal compact representation but also to construct a new space that has structure similar to the original hyperspectral data with least reconstruction error.

The remainder of this paper is organized as follows: Section II explains the concept our proposed method. Section III describes the experimental setup and results. Section IV discusses the comparison with state-of-the-art works. Section V concludes the paper, and finally section VI discusses the possible future directions.

\section{Methodology}
AEs have achieved remarkable results in many practical applications [25][27]. The key concept in an AE is to map the entire input space onto itself. The mapping is found by training a set of encoding and decoding layers using unlabeled samples. Thus, there is no need to consider class information. Once trained, the hidden representation at the innermost encoding layer of an AE can be visualized as a compression of the input space containing key contents or information. DAEs are created by stacking multiple AEs, where the encoders are normally constructed using a greedy layer-wise approach. 

In this work we consider two different strategies to train an AE i.e., segmented and non-segmented AE. Non-segmented AE means that the hyperspectral image cube is processed without any preprocessing or splinting the image cube into parts. Whereas, segmented AE means that the hyperspectral image cube is divided into several parts to train an AE individually on each part of hyperspectral image cube. Segmentation based AE training is proved to incur less computational overhead in terms of time. To train segmented AE, researchers usually tend to segment hyperspectral image cube over spectral domain. Spectral segmentation refers to dividing the wavelengths into several parts to train AE. However, there is no guarantee that the selected bands from particular segments are optimal and preserve the important information. More precisely,  training a wavelength-based segmented AE normally includes uninformative bands where such bands might have high correlation at some wavelength but not at other segment of hyperspectral image cube or may include the redundant bands. Redundant means, the similar bands from different segments which might contain no additional information for band reduction and classification.

To overcome the above-discussed problem, in this work, we propose a spatial information-based hyperspectral image segmentation to train an AE. In this process, we split the hyperspectral image cube based on its pixel locations rather than on spectral or wavelength values. An AE trained on a segment of hyperspectral image cube obtained through spatial information, significantly improves the generalization abilities for both supervised and unsupervised tasks, i.e., classification and clustering. However, although, these segments are obtained by spatial information, the training of an AE on these segments are done based on their spectral values. In other words, this process dose not incorporate the spatial information while training an AE. While selecting the bands from individual segment, the number and position of selected band from each segment should remain the same. Furthermore, in this work we rigorously compare both segmented and non-segmented AEs for both supervised and unsupervised tasks.

To explain our proposed method theoretically, we assume \(X = [x_1,~x_2,~x_3,~\dots,~x_L]^T \in \mathcal{R}^{L*P}\) the lexicographically indexed hyperspectral image composed of \(L\) bands and \(P\) pixels per band. Let us further assume that \(x_l = [x_{l,1},~x_{l,2},~x_{l,3},~\dots ,~x_{l,P}]^T\) is \(l^{th}\) band that corresponds to a wavelength \(\lambda_l\). Furthermore, we also assume a deterministic mapping \(f_\theta\) that transforms an input band \(x_l = x\) into hidden representation \(y\) which is known as affine nonlinear mapping defined as;

\begin{equation}
f_{\theta} (x) = s(W x~+~b)
\end{equation}
where \(\theta = \{W,~b\}\). Here \(b\) is an \(\hat{l}\)-dimensional offset vector and \(W \in (\hat{l}~\times~l)\) weight matrix. The resulting hidden representation \(y\) is then mapped back to the reconstructed vector \(z\) as \(z = g_{\hat{\theta}} (y)\). The decoder mapping is again an affine mapping followed by squashing nonlinearity and can be define as:

\begin{equation}
g_{\hat{\theta}} (y) = \hat{W}~y + \hat{b} ~\simeq  ~s(\hat{W}~y + \hat{b})
\end{equation}

Generally \(z\) can not be interpreted as an exact reconstruction of \(x\) but rather in probabilistic terms as the parameters of the distribution \(p(\mathcal{X}|Z = z)\) that may generate \(x\) with high probability. For this, let us assume \(p(\mathcal{X}|Y = y) = p(\mathcal{X}|Y = g_{\hat{\theta}} (y))\) which yields an associated reconstruction error to be optimized as \(L(x,z) \propto - log~p(x|z)\). 

Since the pixels in hyperspectral space are normalized, the common choice for \(p(x|z)\) and associated loss function \(L(x,z)\) include \(x \in \mathcal{R}^L : \mathcal{X}|z \sim \mathcal{N}(Z, \phi^2~I)\), where \(\mathcal{X}_j|z \sim \mathcal{N}(z_j, \phi^2)\). This yields \(L(x,z) = L_2(x,z)\), where \(L_2(x,z)\) can be defined as \(L_2(x,z) = C(\phi^2) ||x - z||^2\), where \(C(\phi^2)\) denotes a constant that depends only on \(\phi^2\) which can be neglected for the optimization. The training of an AE consists of \textit{minimizing the reconstruction error} that carries the following optimization:

\begin{equation}
\argminB_{\theta, \hat{\theta}} \mathbb{E}_{q^o(\mathcal{X})} [L(\mathcal{X},Z(\mathcal{X})]
\end{equation}
where \(Z(\mathcal{X})\) refers the fact that \(Z\) is a deterministic function of \(\mathcal{X}\) and it can be incurred by the composition of encoding and decoding with the loss function as defined below:

\begin{equation}
\argmaxB_{\theta, \hat{\theta}} \mathbb{E}_{q^o(\mathcal{X})} [log~p (\mathcal{X}|Z = g_{\hat{\theta}} (f_\theta(\mathcal{X})))]
\end{equation}
\begin{equation}
\argmaxB_{\theta, \hat{\theta}} \mathbb{E}_{q^o(\mathcal{X})} [log~p (\mathcal{X}|Z = f_\theta(\mathcal{X});~\hat{\theta})]
\end{equation}

The loss function corresponds to minimizing reconstruction error i.e. maximizing the lower bound on the mutual information between input \(\mathcal{X}\) and the learnt representation \(Y\). Intuitively, the reconstruction criterion itself is unable to guarantee the extraction of useful bands from hyperspectral data as it can lead to the manifest solution i.e. "copying the input". One way to avoid this phenomenon is to constrain on sparse representation. In this work, we rather constrain on sparsity, we change the reconstruction criterion i.e. denoising. By doing so, we used a better representation, better in terms one that can be obtained robustly from a corrupted bands which will be useful for recovering the corresponding clean bands. To this end, two underlying ideas are implicit:

\begin{enumerate}
\item It is expected that a higher level representation should be stable and robust under corruption of the input bands.

\item It is expected that performing the denoising task requires selecting bands that capture useful structure in the input space.
\end{enumerate}

These two objectives have been obtained by first corrupting the initial input \(x\) to \(\hat{x}\) through stochastic mapping \(\hat{x} \sim q \mathcal{D}(\hat{x}|x)\). The corrupted input band is then mapped to a hidden representation \(y = f_\theta(\hat{x}) = s(W~ \hat{x} +~b)\). From this hidden representation we can reconstruct \(z\) as \(z = g_{\hat{\theta}}(y)\). The parameters \(\theta\) and \(\hat{\theta}\) are trained to minimize the average reconstruction error i.e. to have \(z\) as close as possible to the uncorrupted input \(x\). The key difference is that \(z\) is now a deterministic function of \(\hat{x}\) rather than \(x\). The two different reconstruction error methods are discussed below in which the parameters are randomly initialized and then optimized by stochastic gradient descent method.
\begin{enumerate}
\item Cross Entropy Loss with affine and sigmoid decoder is defined as \(L_{\mathcal{H}}(x,z) = \mathcal{H}(\mathfrak{B}(x) ||\mathfrak{B}(z))\).
\item Squared Loss error with affine decoder is defined as \(L_2(x,z) = ||x - z||^2\).
\end{enumerate}

A different corrupted version of input band segment is generated according to \(q\mathcal{D}(\hat{x}|x)\). Denoising AE also minimizes the same reconstruction loss between input band segment \(\mathcal{X}\) and its reconstruction from \(Y\).  Thus, it accounts to maximizing a lower bound on the mutual information between \(\mathcal{X}\) and its representation \(Y\). The main difference is that \(Y\) is obtained by applying deterministic mapping \(f_\theta\) to a corrupted input. It thus forces the learning of a far more clever mapping than the identity (that selects the bands useful for denoising).

\begin{figure}[!ht]
	\centering
	\includegraphics[scale=0.3]{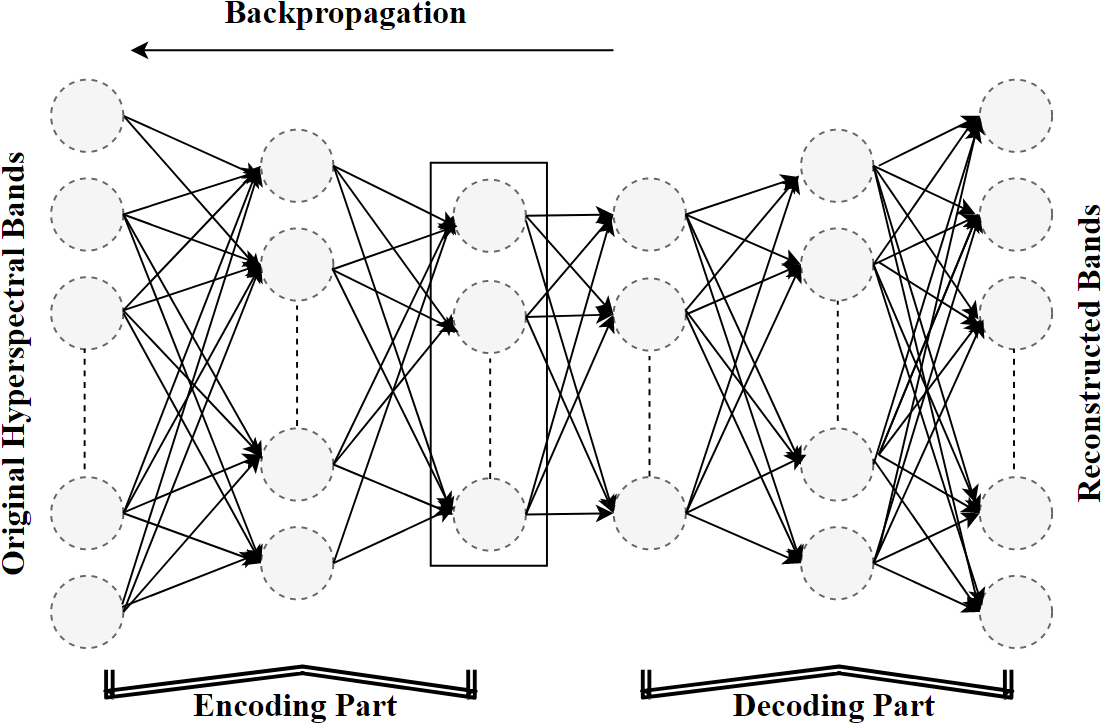}
	\captionsetup{justification=centering}
	\caption{The UDAE network is stacked by two layer AE's.}
	\label{Fig1}
\end{figure}

\begin{figure}[!ht]
	\centering
	\includegraphics[scale=0.67]{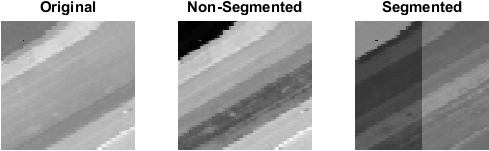}
	\captionsetup{justification=centering}
	\caption{The input band and the output of \textit{UDAE} on single band in reduced dimension for both segmented and non-segmented \textit{UDAE} method.}
	\label{Fig2}
\end{figure}

Figure~\ref{Fig1} shows an instance of the network structure. It includes two encoding and decoding layers. In the encoding layer, output of the first encoding layer acts as the input data for the second encoding layer, and so forth. The inner most representation i.e., the last layer of encoding part is visualized in Figure~\ref{Fig2} as a compression of the input space containing the key contents of hyperspectral bands. Furthermore, Figure~\ref{Fig2} also shows the ability of \textit{UDAE} method to preserve the key information and original structure of hyperspectral data in lower dimensions for both segmented and non-segmented \textit{UDAE}.

\section*{Algorithm-1}
Segmented and non-segmented \textit{UDAE} method in which \(S\) determine the number of segments.
\begin{algorithm}
	\SetAlgoLined
	\KwResult{\(Selected~Bands\); Reconstruction Error = C}
	\textbf{Input}; \(\#Layers\); X; \(\lambda\); \(\psi\) = Noise; \(\Theta\) = Epoch\;
	\textbf{Initialization}; \(\mathcal{L} = r/S; \ A = 1\)\;
	\While{\(j \leq S \leq 2\)}{
		\(\mathcal{X} = X(A:\mathcal{L},:)\)\;
		\For{i = 1 : \(\#Layers\)}{
			[Net, C] = \textbf{AE}(\(\mathcal{X}\), Layers(i), \(\lambda\), \(\psi\) , \(\Theta\))\;
			Model\{i\}.Bias = Net\{1\}.Bias\;
			Model\{i\}.W = Net\{1\}.W\;
			\(C_j\) = [\(C_{j}\) C]\;
		}
		\For{i = 1: \(\#Layers\)}{
			Model\{\(\#Layers\) + i\}.W = Model\{\(\#Layers\) - i + 1\}.W'\;
			\eIf{i $\neq$ \(\#Layers\)}{
				Model\{\(\#Layers\)\}.Bias = Model\{\(\#Layers\) -i\}.Bias\;
			}{
				Model\{\(\#Layers\)\}.Bias = zeros(1, size(\(\mathcal{X}\), 2))\;
			}
		}
		\textit{Fine-Tuning}\;
		\(Model\{j\}\) = \textbf{BP}(\(Model\{j\}\), \(\mathcal{X}\), \(\mathcal{X}\), \(\lambda\), \(\psi\), \(\Theta\))\;
		\textit{BP stands for batch-based backpropagation method}\;
		\textit{Final Model for Band Selection}\;
		\(Selected~Bands_j\) = \textbf{ActivativeFun}(\(Model\{j\}\), \(\mathcal{X}\))\;
		A = \(\mathcal{L}\)\;
		\(\mathcal{L}\) = (\(\mathcal{L}\) +1) + (\(\mathcal{L}\)-2)\;
	}
\end{algorithm}
\vspace{-0.5cm}
\section*{Algorithm-2}
Training of Autoencoder (\textbf{AE Function}).
\begin{algorithm}
	\SetAlgoLined
	\KwResult{Net; C;}
	\textbf{Input}; \(\#Layers\); \(\mathcal{X}\); \(\lambda\); \(\psi\); \(\Theta\)\;
	\textbf{Initialization}; Net; \(\mathcal{P}\)\; 
	\For{i = \(\#Layers\) - 1}{
		Net\{i\}.W = randn(Layers(i-1), layers(i))\(\lambda\)\;
		Net\{i\}.Bias = zeros(1, Layers(i))\;
	}
	Net\{\(\#Layers\)\}.Bias = zeros(1, \(\mathcal{P}\))\;
	Net\{\(\#Layers\)\}.W = randn(Layers(end), \(\mathcal{P}\))\(\lambda\)\;
	[Net, C] = \textbf{BP}(Network, \(\mathcal{X}\), \(\mathcal{X}\), \(\lambda\), \(\psi\), \(\Theta\))\;
\end{algorithm}

Machine learning methods often suffer from hyperspectral image classification due to the curse of dimensionality. This is because in the hyperspectral space, each sample consists of hundreds of potential bands. Eventually evaluating every sample from each band can affect not only the classifiers efficiency but also may reduce the classification power. Unlike the traditional machine learning methods, our proposed \textit{UDAE} method selects only those bands for classification that are known to improve the predictive power of the model and potentially improve the execution time because irrelevant bands does not need to be preserved. In this work, we used several external classifiers to test the efficiency of \textit{UDAE} method. These classification methods include support vector machine (SVM) [28][31], k-nearest neighbors (kNN) [32] and ensemble learning classifiers (Ada-boost (AB), gentle-boost (GB), logi-boost (LB) and Bag) [2], [33][36].

\section{Experimental Settings and Results}

\subsection{Experimental Datasets}
\begin{figure}[!ht]
\centering
\includegraphics[scale=0.28]{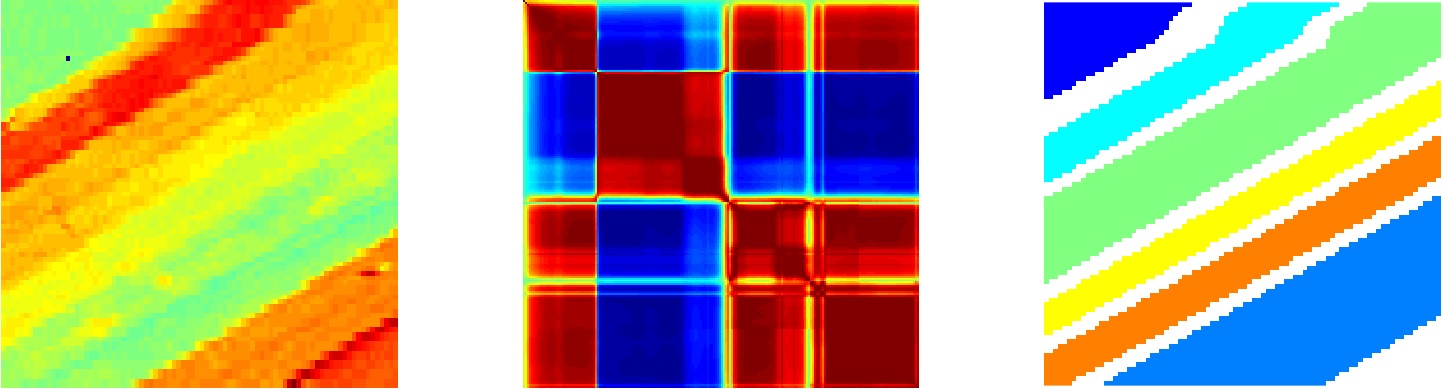}
\captionsetup{justification=centering}
\caption{Composition of Spectral Bands, Correlation among Bands and Ground Truth Maps for Salinas-A Dataset.}
\label{fig.1}
\end{figure}

To this end, we use a publicly available real hyperspectral dataset [37][38] to evaluate the performance of \textit{UDAE} method. The experimental dataset consists of a sub-scene of 224 spectral bands with a spatial resolution of 3.7m. Some bands were water absorption and were removed prior to the analyses. The removed bands are 108-112, 154-167 and 224. The full Salinas scene is covered with $512 \times 217$ samples per band and contains vegetables, bare, soils and vineyard field. Salinas-A ground truth consists of 6 classes and $86 \times 83$ samples per band. The selected samples are located in the original scene at 591-676 and 158-240. The selected dataset class description is provided in the Table~\ref{Tab.1} and the ground truth map is presented in Figure~\ref{fig.1}.

\begin{table}[!ht]
	\captionsetup{justification=centering}
	\centering
	\caption{Number of Samples per Class.}
	\begin{tabular}{lcc}
		\hline
		\textbf{Class} & \textbf{Class Name}    & \textbf{Number of Samples} \\ \hline
		1 & Brocoli green weeds 1 & 391 \\ \hline 
		2 & Corn senesced green weeds	& 1343 \\ \hline
		3 & Lettuce romaine 4wk	& 616 \\ \hline
		4 & Lettuce romaine 5wk	& 1525\\ \hline
		5 & Lettuce romaine 6wk	& 674 \\ \hline
		6 & Lettuce romaine 7wk & 799\\ \hline
		\label{Tab.1}
	\end{tabular}
\end{table}

\subsection{Experimental Settings}
The purpose of this work is to investigate the efficient representation of hyperspectral data using \textit{UDAE} method. Efficient representation refers to the representation in a lower dimensional space from which it is possible to reconstruct the original space with least reconstruction error. In \textit{UDAE} method, the last activation function is set as a linear function because of spectral target values. In this process, we observe two different learning models: 1): batch-Model: the gradient is computed as an average for a set of spectral vectors, 2): stochastic-Model: the net model is adapted to one spectral vector at a time. While implementing \textit{UDAE}, we found that there is a significant difference in speed of going through an epoch in the learning model which leads to conclude a solution between the extremes by using tiny batches. The learning rate is determined by hit and trial based on an error just to speed up the convergence rate of learning model. Prior to analyzing the learning rate, we normalize the data just to improve the convergence of the learning model by translating each hyperspectral band to obtain zero mean.

The DAE network configuration consists of several parameters such as the number of hidden layers, number of neurons in each hidden layer and learning rate to control the reconstruction and fine-tuning process and a maximum number of iterations to train each layer. The number of hidden layers are selected in the range from 4-6 and the number of hidden neurons are chosen as a set of \(\{Number-of-bands, \ 3*Number-of-bands-to-select, \ 2*Number-of-bands-to-select, \ Number-of-bands-to-select\}\) in case of 4 hidden layers. The learning parameters are selected from the range [0.001 0.1] and the maximum number of iterations for training and fine-tuning are randomly set between \(\{200-250\}\). All these parameters are carefully tuned for the given experimental results. The optimal selection of these parameters are obtained according to the optimal classification performance. Carefully tuned \textit{UDAE} method is tested using three different classifiers as explained in Section II-B. The important parameters to train these classifiers are as follows: SVM classifier is trained using 'RBF' kernel function. kNN classifier is trained using the range from \(\{1-20\}\) nearest neighbors. The ensemble learning for both AB, GB, LB and BAG are trained using tree-based model with \(\{1-100\}\) and \(\{1-30\}\) number of trees respectively.

\subsection{Experimental Results}
The Kappa (\(\kappa\)) coefficient and overall accuracies are considered as the evaluation metrics since these are widely used in existing works [39][40]. The \(\kappa\) coefficient is obtained by the following expressions:

\begin{eqnarray}
\kappa = \frac{n \sum_k\Psi_k - \sum_k\sigma_k\phi_k}{n^2 - \sum_k\sigma_k\phi_k}
\end{eqnarray}
where \textit{N} is the total number of samples, \(\Psi_k \) represents the number of correctly predicted samples in the given class. \(\sum_k \Psi_k \) is the sum of the number of correctly predicted samples. \(\sigma_k \) is the actual number of samples belonging to the given class and \(\phi_k \) is the number of samples that have been correctly predicted into the given class [39].

\begin{figure}[!ht]
	\centering
	\includegraphics[scale=0.5]{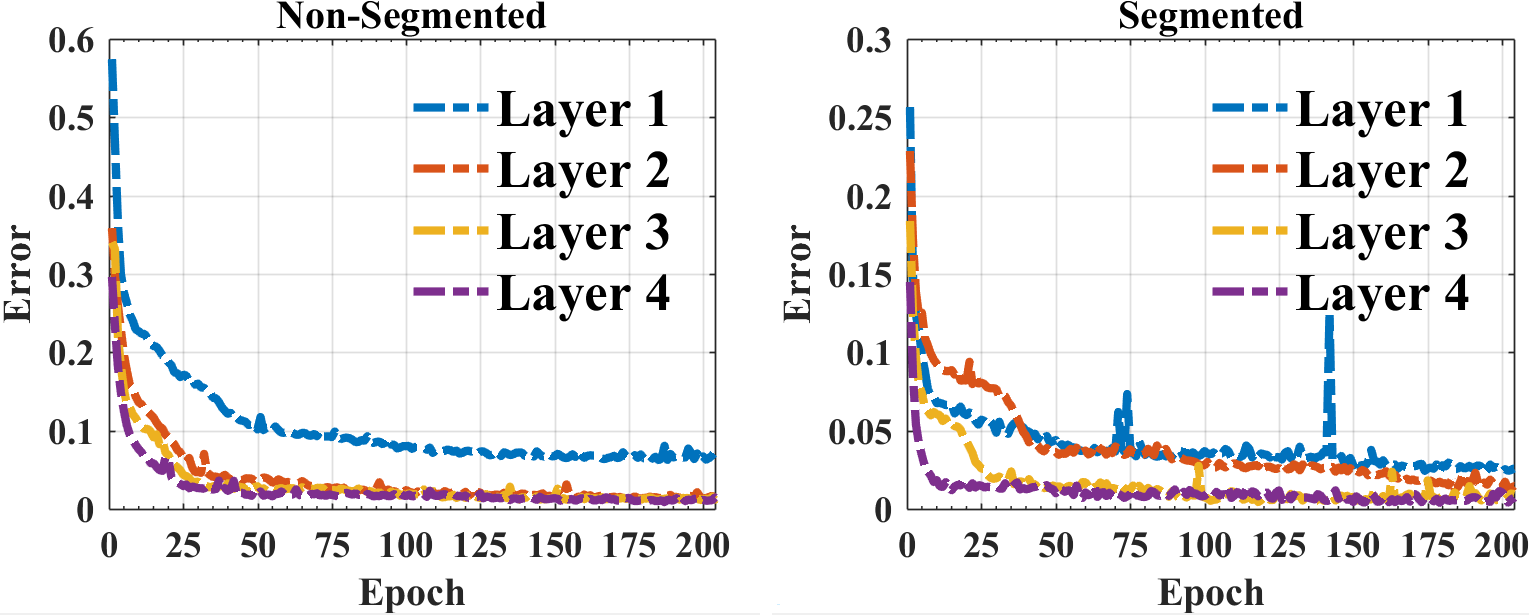}
	\captionsetup{justification=centering}
	\caption{Reconstruction error on each layer against the number of iterations epochs for both segmented and non-segmented \textit{UDAE}.}
	\label{fig.2}
\end{figure}
In our first experiment, we provide sensitivity analysis of reconstruction of the actual data from the DAE network, based on the provided settings as explained in section III-B. Figure~\ref{fig.2} shows the results of reconstruction error on each layer with respect to the number of epochs on each layer for both Segmented and non-segmented \textit{UDAE}.

\begin{figure}[!ht]
	\centering
	\includegraphics[scale=0.35]{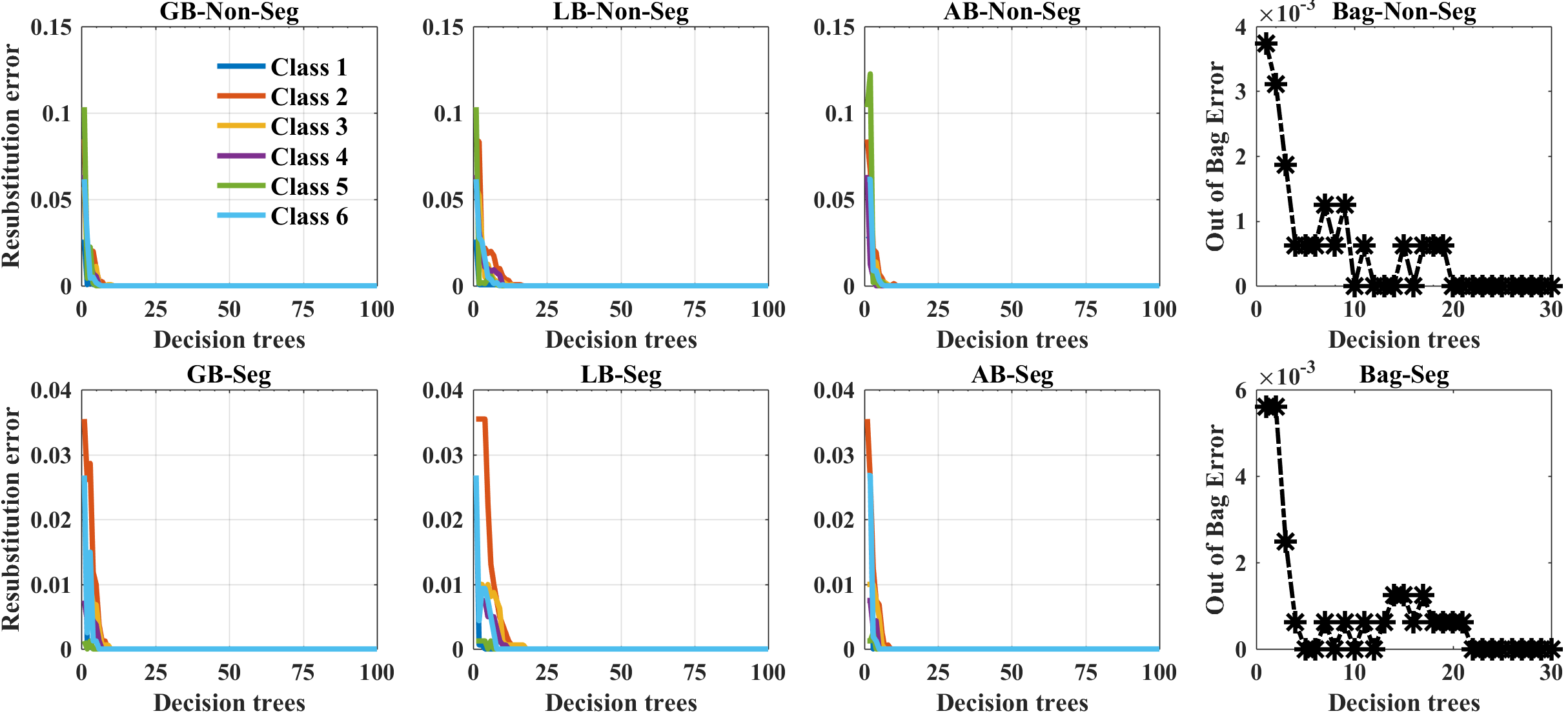}
	\captionsetup{justification=centering}
	\caption{Resubstitution and Out-of-Bag errors for both segmented and non-segmented \textit{UDAE}.}
	\label{fig.3}
\end{figure}
\begin{figure}[!ht]
	\centering
	\includegraphics[scale=0.55]{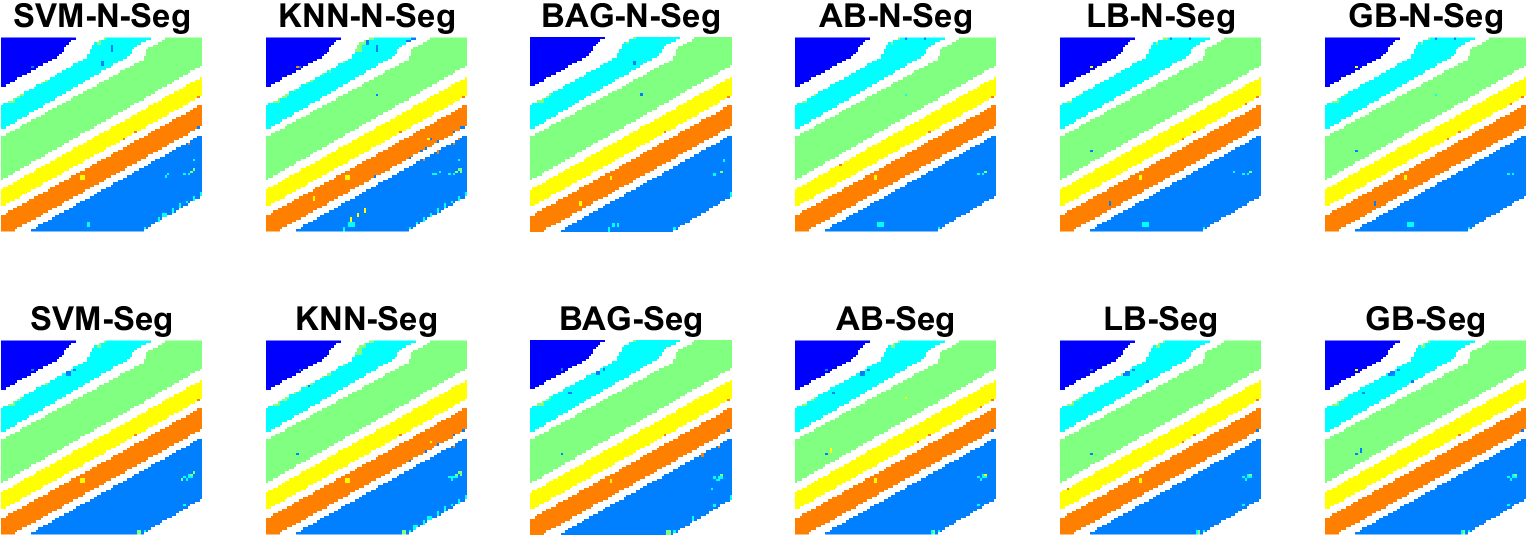}
	\captionsetup{justification=centering}
	\caption{Ground Truth Maps Obtained by SVM, kNN, GB, LB, AB and Bag Classifiers for both segmented and non-segmented \textit{UDAE}, respectively.}
	\label{fig.4}
\end{figure}
\begin{figure}[!ht]
	\centering
	\includegraphics[scale=0.4]{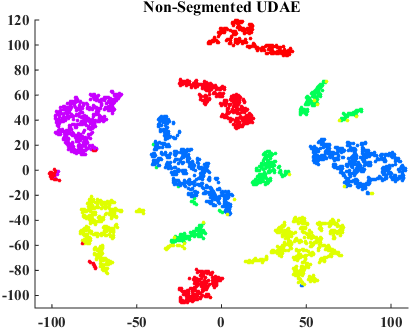}
   	\includegraphics[scale=0.4]{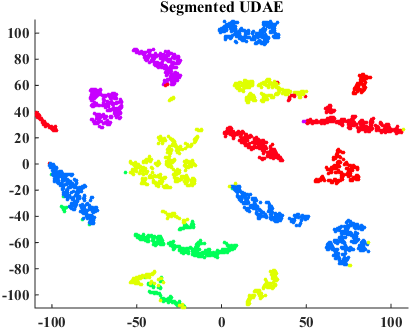}
	\captionsetup{justification=centering}
	\caption{Clusters learned by the hidden representation of both segmented and non-segmented \textit{UDAE}, respectively.}
	\label{fig.5}
\end{figure}

In the second experiment, we analyze the behavior of different ensemble learning classifiers based on re-substitution and out-of-bag error against different numbers of decision trees. Figure~\ref{fig.3} shows the results of re-substitution and out-of-bag error for GB, LB, AB and Bag classifiers respectively. Figure~\ref{fig.4} shows the visual classification results of the six classifiers, i.e., SVM, kNN, GB, LB, AB and Bag respectively. Figure~\ref{fig.5} shows the visual clustering results obtained from both segmented and non-segmented \textit{UDAE} inner most hidden representation. In this work, we used t-Distributed Stochastic Neighbor Embedding (t-SNE) with pair-wise distance metric method to visualize the clusters in 2-Dimensional space. From results, we observe that the non-segmented \textit{UDAE} method performs better in terms of defining the clusters than segmented \textit{UDAE}. Segmented \textit{UDAE} breaks the similar clusters into several parts which helps to obtain better classification and generalization performance in computationally efficient fashion than the non-segmented \textit{UDAE}. The obtained results demonstrate the acceptable generalization capability for the \textit{UDAE} method for all these three types of classifiers.

\section{Comparison with State of the Art}
The proposed \textit{UDAE} method is compared with state-of-the-art such as Principle Component Analysis (PCA) [4] and Laplacian Eigenmaps (LEM's) [41]. Both methods are unsupervised, Laplacian Eigenmaps method is non-linear unsupervised and PCA is linear mapping method. To maintain the consistency between these methods, 50 most informative bands were extracted empirically by each method. After carefully extracting the bands, SVM, kNN, GB, LB, AB and Bag classifiers were used to classify the extracted bands.

Table~\ref{Tab.2} and \ref{Tab.3} show the classification accuracy obtained by SVM, kNN, GB, LB, AB and Bag classifiers respectively. These results shows that the proposed \textit{UDAE} method outperforms other well-known state-of-the-art non-linear BR methods in terms of kappa coefficient and overall accuracy. From results we conclude that the instance-based classifiers do not lead to better performance as compared to the ensemble learning classifiers because no spatial structure of hyperspectral data have been considered. It also might be the case that the training samples were randomly selected that lead to the class imbalance issue. For example the rocoli green weeds-1 class has 391 samples as compared to the Lettuce romaine 5wk and Corn senesced green weeds which have 1525 and 1343 samples respectively. In our experiments, we randomly chose 30\% of samples per class for training model and the remaining samples were used as a test data.

\begin{table}[!ht]
	\captionsetup{justification=centering}
	\centering
	\caption{kappa Coefficient Comparisons.}
	\begin{tabular}{lcccc}
		\hline
		\textbf{Classifiers} & \textbf{Seg-UDAE} & \textbf{Non-Seg-UDAE} & \textbf{PCA} & \textbf{LEM} \\ \hline
		SVM  & 0.9849 & 0.9883 & 0.6967 & 0.8125 \\ \hline
		kNN  & 0.9742 & 0.9799 & 0.7751 & 0.4476  \\ \hline
		AdaBoost  & 0.9923 & 0.9873 & 0.7725 & 0.4758 \\ \hline
		GentleBoost  & 0.9903 & 0.9879 & 0.9157 & 0.9439 \\ \hline
		LogiBoost  & 0.9903 & 0.9866 & 0.8905 & 0.9533 \\ \hline
		BAG  & 0.9903 & 0.9863 & 0.8929 & 0.9716 \\ \hline
       \label{Tab.2}
	\end{tabular}
\end{table}

\begin{table}[!ht]
	\captionsetup{justification=centering}
	\centering
	\caption{Overall Classification Accuracy Comparison.}
	\begin{tabular}{lcccc}
		\hline
		\textbf{Classifiers} & \textbf{Seg-UDAE} & \textbf{Non-Seg-UDAE} & \textbf{PCA} & \textbf{LEM} \\ \hline
		SVM  & 0.9906 & 0.9879 & 0.7656 & 0.8518 \\ \hline
		kNN  & 0.9839 & 0.9794 & 0.7753 & 0.4477 \\ \hline
		AdaBoost  & 0.9898 & 0.9939 & 0.8209 & 0.6075 \\ \hline
		GentleBoost  & 0.9893 & 0.9922 & 0.9135 & 0.9547 \\ \hline
		LogiBoost  & 0.9893 & 0.9922 &  0.9325 & 0.9637 \\ \hline
		BAG  & 0.9890 & 0.9922 & 0.9146 & 0.9772 \\ \hline
       \label{Tab.3}
	\end{tabular}
\end{table}

\section{Conclusion and Future Directions}
In this work a non-linear DAE-based hyperspectral BR method is proposed where \textit{(UDAE)} pre-train and fine-tune the network in an unsupervised fashion on each segment of original hyperspectral data where the segmented \textit{UDAE} is used otherwise hyperspectral has been processed without segmentation.  Furthermore the performance of \textit{UDAE} BR method has been tested by several classifiers where all these classification methods are tested in semi-supervised manner. From results we observe that the proposed method outperforms while using ensemble learning classifier. We carefully tuned \textit{UDAE} method which also provides better classification and clustering results than several popular BR methods.

In the current work, we only considered a subset of Salinas dataset inside the main text due to the page limit but the other high dimensional datasets description and experimental results are attached as supplementary material. These datasets include Pavia University, Pavia Centre, Salinas full scene, Botswana and Kennedy Space Dataset. All of these datasets are Earth Observation images taken from different airborne sensors such as, Airborne Visible/Infrared Imaging Spectrometer (AVIRIS) and Reflective Optics System Imaging Spectrometer (ROSIS) or Satellite such as NASA EO-1.

Currently, we only explored the possibilities to use spectral information for band reduction; in this direction there are several methods that can be used to fuse the spatial information. These methods include morphological or spatial coordinate information and relations between spatial adjacent pixels. Furthermore, in our current work, we only consider a heuristic approach based on global metric to find the optimal number of bands and to measure the performance at inner most hidden representation of our proposed \textit{UDAE} method. However, in future, we intend to incorporate  local metrics such as structure similarity index measure (SSIM) and multi-scale structure similarity index measure (MS-SSIM) to further validate the quality of hidden representation. Furthermore,  we plan to incorporate the regularization method to optimize the number of bands to be extracted systematically with several types of AEs e.g. sparse, regularized and variational AEs.

\section*{References}
\begin{enumerate}[ {[}1{]} ]
\scriptsize
	\item Fan Li, Linlin Xu, P. Siva, et al., "Hyperspectral Image Classification with Limited Labeled Training Samples Using Enhanced Ensemble Learning and Conditional Random Fields", IEEE Journal of Selected Topics in Applied Earth Observations and Remote Sensing, vol. 8(6), pp. 2427-2438, 2015.
	
	\item M. Ahmad, A. M. Khan, and R. Hussain, "Graph-based Spatial-Spectral Feature Learning for Hyperspectral Image Classification", IET image processing, vol. 11(12), p. 1310–1316, 2017. 
	
	\item M. Ahmad, I. U. Haq, Q. Mushtaq, et al., "A new statistical approach for band clustering and band selection using K-means clustering", International Journal of Engineering and Technology, vol. 3(6), pp. 606-614, 2011.
	
	\item C. Rodarmel and J. Shan, "Principal Component Analysis for Hyperspectral Image Classification", Surveying and Land Information Systems, vol. 62(2), pp. 115-123, 2002.
	
	\item M. Fauvel, J. Chanussot, and J. A. Benediktsson, "Kernel Principal Component Analysis for the Classification of Hyperspectral Remote Sensing Data over Urban Areas", EURASIP Journal on Advances in Signal Processing, vol. 2009, 14 pages, doi:10.1155/2009/783194, 2009.
	
	\item T. V. Bandos, L. Bruzzone, and G. Camps-Valls, "Classification of Hyperspectral Images With Regularized Linear Discriminant Analysis", IEEE Transactions on Geoscience and Remote Sensing, vol. 47(3), pp. 862–873, 2009.
	
	\item M. Ahmad, A. M. Khan, J. A. Brown, et al., "Gait fingerprinting-based user identification on smartphones", In proc. of IEEE International joint conference on Neural Networks (IEEE IJCNN) in conjunction with World Congress on Computational Intelligence (IEEE WCCI), pp. 2161-4407, 2016.
	
    \item M. H. Siddiqi,  Seok-Won Lee and A. M. Khan, "Weed Image Classification using Wavelet Transform,
Stepwise Linear Discriminant Analysis, and Support Vector
Machines for an Automatic Spray Control System", Journal of Information Science and Engineering, vol. 30, pp. 1253-1270, 2014.
	\item T. R. D. Saputri, A. M. Khan and Seok-Won Lee, "User-independent activity recognition via three-stage GA-based feature selection", International Journal of Distributed Sensor Networks, vol. 10, no. 3, pp. 706287, 2014.    
    \item W. Li, S. Prasad, J. E. Fowler, et al., "Locality-Preserving Discriminant Analysis in Kernel-Induced Feature Spaces for Hyperspectral Image Classification", IEEE Geoscience and Remote Sensing Letters, vol. 8(5), pp. 894-898,  2011.
	
	\item Y. Yuan, G. Zhu, and Q. Wang, "Hyperspectral band selection by multitask sparsity pursuit", IEEE Transaction on Geoscience and Remote Sensing, vol. 53(2), pp. 631-644, 2015.
	
	\item N. Wadstromer and D. Gustafsson, "Non-Linear Hyperspectral Subspace Mapping using Stacked Autoencoder", The 29th Annual Workshop of the Swedish Artificial Intelligence Society (SAIS), pp. 2-3, 2016.
	
	\item M. Ahmad, I. U. Haq, and Q. Mushtaq, "AIK Method for Band Clustering Using Statistics of Correlation and Dispersion Matrix",  In proc. of International Conference on Information Communication and Management (IPCSIT), vol.16, pp. 114-118, 2011.
	
	\item M. Ahmad, A. M. Khan, R. Hussain, et al., "Unsupervised geometrical feature learning from hyperspectral data", In proc. of IEEE  Symposium Series on Computational Intelligence (SSCI), pp. 1-6, 2016. 
	
	\item C.I. Chang and Q. Du, "Estimation of number of spectrally distinct signal sources in hyperspectral imagery", IEEE Transaction on Geoscience and Remote Sensing, vol. 42(3), pp. 608-619, 2004.
	
	\item H. Su, Qian Du, G. Chen, et al., "Optimized hyperspectral band selection using particle swarm optimization", IEEE Journal of Selected Topics in Applied Earth Observations and Remote Sensing, vol. 7(6), pp. 2659-2670, 2014.
	 
	\item H. Su, B. Yong, and Q. Du, "Hyperspectral Band Selection Using Improved Firefly Algorithm", IEEE Geoscience and  Remote Sensing Letters, vol. 13(1), pp. 68-72, 2016. 
	
	\item A. Ifarraguerri, "Visual method for spectral band selection", IEEE Geoscience and  Remote Sensing Letters, vol. 1(2), pp. 101-106, 2004. 
	
	\item C.I. Chang, "Hyperspectral Imaging: Techniques for Spectral Detection and Classification", New York, NY, USA: Kluwer, ch. 2, pp. 15-35, 2003. 
	
	\item A. Martinez-Uso, F. Pla, J. M. Sotoca, et al., "Clustering-based hyperspectral band selection using information measures", IEEE Geoscience and Remote Sensing Letters, vol. 45(12), pp. 4158-4171, 2007. 
	
	\item H. Yang, Qian Du, Hongjun Su, et al., "An efficient method for supervised hyperspectral band selection", IEEE Geoscience and Remote Sensing Letters, vol. 8(1), pp. 138-142, 2011. 
	
	\item R. Nakamura, L. M. G. Fonseca, J. A. D. Santos, et al., "Nature-inspired framework for hyperspectral band selection", IEEE Transaction on Geoscience and Remote Sensing, vol. 52(4), pp. 2126-2137, 2014. 
	
	\item X. Yang, "Nature-Inspired Metaheuristic Algorithms", Bristol, U.K.: Luniver Press, 2008.
	
	\item X. Yang and X. He, "Firefly algorithm: Recent advances and applications", International Journal of Swarm Intell., vol. 1(1), pp. 36-50, 2013.
	
	\item D. Gustafsson, H. Petersson, and M. Enstedt, "Deep learning: Concepts and selected applications", Technical Report FOID-0701-SE, Swedish Defence Research Agency (FOI), Sweden, 2015.
	
	\item X. W. Chen and X. Lin, "Big data deep learning: Challenges and perspectives", Access IEEE, pp. 514-525, 2014.
	
	\item Y. Bengio, "Learning deep architectures for AI", Foundations and Trends in Machine Learning, vol. 2(1), pp. 1-127, 2009.
	
	\item T. Hastie, R. Tibshirani, and J. Friedman, "The Elements of Statistical Learning", second edition. New York: Springer, 2008.
	
	\item N. Christianini, and J. Shawe-Taylor, "An Introduction to Support Vector Machines and Other Kernel-Based Learning Methods", Cambridge, UK: Cambridge University Press, 2000.
	
	\item R. E. Fan, P. H. Chen, and C. J. Lin, "Working set selection using second order information for training support vector machines", Journal of Machine Learning Research, vol 6, pp. 1889–1918, 2005.
	
	\item V. Kecman, T. M. Huang, and M. Vogt, "Iterative Single Data Algorithm for Training Kernel Machines from Huge Data Sets: Theory and Performance", In Support Vector Machines: Theory and Applications. Edited by Lipo Wang, pp. 255–274, 2005.
	
	\item J. H. Friedman, J. Bentely, and R. A. Finkel, "An Algorithm for Finding Best Matches in Logarithmic Expected Time", ACM Transactions on Mathematical Software, vol. 3(3), pp. 209–226, 1977.
	
	\item L. Breiman, "Bagging Predictors", Machine Learning. vol. 26, pp. 123-140, 1996.
	
	\item L. Breiman, "Random Forests", Machine Learning. vol. 45, pp. 5-32, 2001.
	
	\item P. Bartlett, Y. Freund, W. S. Lee, et al., "Boosting the margin: A new explanation for the effectiveness of voting methods", Annals of Statistics, vol. 26(5), pp. 1651-1686, 1998.
	
	\item M. Warmuth, J. Liao, and G. Ratsch, "Totally corrective boosting algorithms that maximize the margin", In proc. of 23rd International Conference on Machine Learning, ACM, New York, pp. 1001-1008, 2006.
	
	\item http://www.ehu.eus/ccwintco/index.php?title=GIC-experimental-databases, access year 2017.
	
	\item M. F. Baumgardner, L. L. Biehl, and D. A. Landgrebe, "220 Band AVIRIS Hyperspectral Image Data Set: June 12, 1992 Indian Pine Test Site 3, Purdue University Research Repository", doi:10.4231/R7RX991C, https://purr.purdue.edu/publications/1947/1, 2015, access year 2017.
	
	\item M. Ahmad, S. Protasov, A. M. Khan, et al., "Fuzziness-based Active Learning Framework to Enhance Hyperspectral Image Classification Performance for Discriminative and Generative Classifiers", PlosOne, DOI: 10.1371/journal.pone.0188996, vol. 13(1), 2018.
	
	\item Y. Gu, C. Wang, Di You, et al., "Representative Multiple Kernel Learning for Classification in Hyperspectral Imagery", IEEE Transactions on Geosciences and Remote Sensing, vol. 50(7), pp. 2852-2865, 2012.
	
	\item D. C. Nathan, D. B. Start, and S. E. Chew, "Modularity versus Laplacian Eigenmaps for dimensionality reduction and classification of hyperspectral imagery", In proc. of 6th workshop on Hyperspectral Image and Signal Processing: Evolution in Remote Sensing (WHISPERS), 2014. 	
\end{enumerate}

\section*{\textbf{Supplementary Material}}
The performance of our proposed \textit{UDAE} method have been tested on several benchmark hyperspectral datasets. These datasets include Pavia Centre, Pavia University, Salinas scene, Kennedy Space Center and Botswana Datasets. All of these datasets are Earth Observation images taken from different airborne sensors such as, Airborne Visible/Infrared Imaging Spectrometer (AVIRIS) and Reflective Optics System Imaging Spectrometer (ROSIS) or Satellite such as NASA EO-1. All these datasets are freely available at [35].

\begin{figure}[!ht]
	\centering
	\begin{subfigure}{0.15\textwidth}
		\centering
		\includegraphics[scale=0.23]{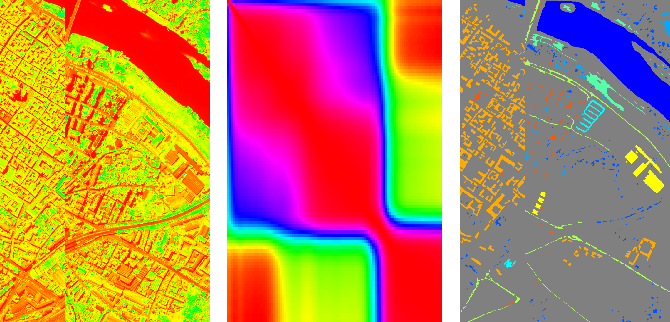}
		\caption{Pavia Centre}
	\end{subfigure}%
	\begin{subfigure}{0.45\textwidth}
		\centering
		\includegraphics[scale=0.2]{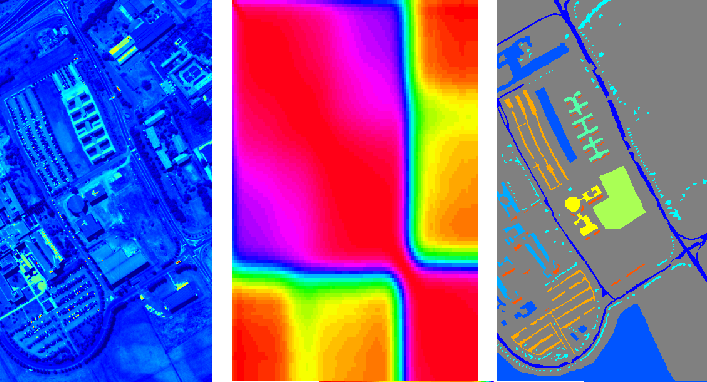}
		\caption{Pavia University}
	\end{subfigure}
	\caption{Composition of Spectral Bands, Correlation among Bands and Ground Truth Maps for Pavia Centre and Pavia University Dataset.}
	\label{Fig.8}
\end{figure}

Pavia Centre (PC) and Pavia University (PU) dataset was acquired by the ROSIS sensor during a flight campaign over Pavia, Northern Italy. PC and PU dataset consist of 102 and 103 spectral bands, where each band is consist of 1096*1096 and 610*610 pixels per band. Some of the pixels in each band contain no information and have to be discarded before the analysis. The discarded pixels in the dataset can be seen as abroad black strips. PC and PU dataset acquired with geometric resolution of 1.3 meters. PC and PU dataset ground truths differentiate nine classes. PC and PU datasets class description is provided in the Table~\ref{Tab.4} and the ground truth map is presented in Fig.~\ref{Fig.8}.

\begin{table}[!ht]
\centering
\tiny
\caption{Class Names and Total Number of Samples for Pavia Centre and Pavia University Datasets.}
\begin{tabular}[t]{cc} 
\multicolumn{2}{c}{\textbf{Pavia Centre.}} \\ \hline
     \textbf{Class Names} & \textbf{Samples} \\ \hline
		Shadows	 & 476 \\ \hline
		Bitumen	 & 808 \\ \hline
        Self-Blocking Bricks & 808 \\ \hline
		Asphalt	 & 816 \\ \hline
        Trees	 & 820 \\ \hline
        Bare Soil	 & 820 \\ \hline
        Water	 & 824 \\ \hline
        Meadows	 & 824 \\ \hline
		Tiles	 & 1260 \\ \hline
    \end{tabular}%
    \quad \ \ \ \ \ \ \ 
    \begin{tabular}[t]{cc}
    \multicolumn{2}{c}{\textbf{Pavia University.}} \\ \hline
     \textbf{Class Names} & \textbf{Samples} \\ \hline
		Shadows	& 947 \\ \hline
        Bitumen	& 1330 \\ \hline
        Painted metal sheets	& 1345 \\ \hline
        Gravel	& 2099 \\ \hline
        Trees	& 3064 \\ \hline
        Self-Blocking Bricks	& 3682 \\ \hline
        Bare Soil	& 5029 \\  \hline
        Asphalt	& 6631 \\ \hline
        Meadows	& 18649 \\ \hline
        \label{Tab.4}
    \end{tabular}
\end{table}

\begin{figure}[!ht]
	\centering
	\begin{subfigure}{0.15\textwidth}
		\centering
		\includegraphics[scale=0.23]{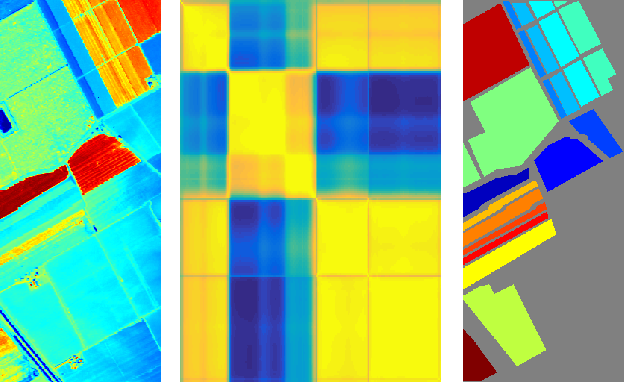}
		\caption{Salinas}
	\end{subfigure}%
	\begin{subfigure}{0.45\textwidth}
		\centering
		\includegraphics[scale=0.2]{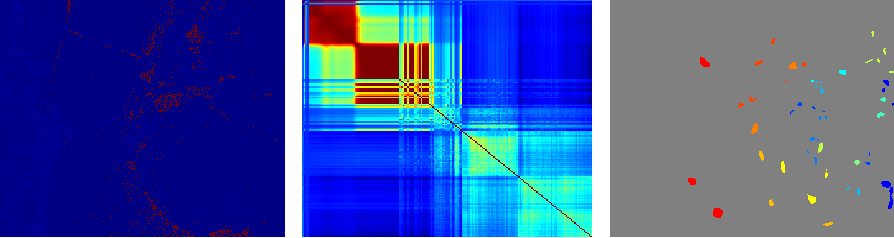}
		\caption{Kennedy Space Centre}
	\end{subfigure}
	\caption{Composition of Spectral Bands, Correlation among Bands and Ground Truth Maps for Salinas and Kennedy Space Centre Datasets.}
	\label{Fig.9}
\end{figure}

Salinas scene was collected by the AVIRIS sensor over Salinas Valley, California Salinas scene is characterized by 3.7 meter per pixel high spatial resolution. Salinas scene is composed by 224 spectral bands. The Salinas area covered comprises 512 lines by 217 samples, i.e. each band is consist of 512*217 pixels. In Salinas dataset, 20 bands were water absorption and have to be discarded prior to the experiments. The discarded bands are 108-112, 154-167 and 224. The Salinas image was available only as at-sensor radiance data. It includes vegetables, bare soils, and vineyard fields. Salinas ground truth contains 16 classes. The selected dataset class description is provided in the Table~\ref{Tab.5} and the ground truth map is presented in Fig.~\ref{Fig.9}.

\begin{table}[!ht]
\centering
\tiny
\captionsetup{justification=centering}
\centering
\caption{Class Names and Total Number of Samples for Salinas and Kennedy Space  Datasets.}
\begin{tabular}{cc}
\multicolumn{2}{c}{\textbf{Salinas Dataset.}} \\ \hline
\textbf{Class Name}    & \textbf{Number of Samples} \\ \hline
		Brocoli green weeds 1 & 2009 \\ \hline 
		Brocoli green weeds 2 & 3726 \\ \hline
		Fallow & 1976 \\ \hline
		Fallow rough plow & 1394\\ \hline
		Fallow smooth & 2678 \\ \hline
		Stubble & 3959\\ \hline
        Celery	& 3579\\ \hline
		Grapes untrained	& 11271 \\ \hline
		Soil vinyard develop & 6203\\ \hline
        Corn senesced green weeds & 3278 \\ \hline
		Lettuce romaine 4wk & 1068\\ \hline
		Lettuce romaine 5wk & 1927 \\ \hline
		Lettuce romaine 6wk & 916\\ \hline
        Lettuce romaine 7wk	& 1070\\ \hline
		Vinyard untrained	& 7268 \\ \hline
		Vinyard vertical trellis & 1807\\ \hline
\end{tabular}
\quad
\begin{tabular}{cc}
\multicolumn{2}{c}{\textbf{Kennedy Space Centre Dataset.}} \\ \hline
\textbf{Class Name}    & \textbf{Number of Samples} \\ \hline
Scrub  & 761\\ \hline
Willow Swamp & 243\\ \hline
CP hammock & 256\\ \hline
CP/Oak & 252\\ \hline
Slash Pine & 161\\ \hline
Oak/Broadleaf & 229\\ \hline
Hardwood Swamp & 105\\ \hline
Graminoid Marsh & 431\\ \hline
Spartina Marsh & 520\\ \hline
Cattail Marsh & 404\\ \hline
Salt Marsh & 419\\ \hline
Mud Flats & 503\\ \hline
Water & 927\\ \hline
\label{Tab.5}
\end{tabular}
\end{table}

Kennedy Space dataset was acquired on March 23, 1996, by NASA AVIRIS instrument over the Kennedy Space Center (KSC), Florida. AVIRIS acquires KSC data in 224 spectral bands of 10nm width with center wavelengths from 400-2500 nm. The KSC data acquired from an altitude of approximately 20km and have a spatial resolution of 18 m. After removing water absorption and low SNR bands, 176 remaining bands were used for the analysis. Training data were selected using land cover maps derived from color infrared photography provided by the KSC and Landsat Thematic Mapper (TM) imagery. The vegetation classification scheme was developed by KSC personnel in an effort to define functional types that are discernible at the spatial resolution of Landsat and these AVIRIS data. Discrimination of land cover for this environment is difficult due to the similarity of spectral signatures for certain vegetation types. For classification purposes, 13 classes representing the various land cover types that occur in this environment were defined for the site. The selected dataset class description is provided in the Table~\ref{Tab.5} and the ground truth map is presented in Fig.~\ref{Fig.9}.

The NASA EO-1 satellite acquired a sequence of data over the Okavango Delta, Botswana in 2001-2004. The Hyperion sensor on EO-1 acquires data at 30 m pixels resolution over a 7.7 km strip in 242 bands covering the 400-2500 nm portion of the spectrum in 10 nm windows. Preprocessing of the data was performed by the UT center for Space Research to mitigate the effects of bad detectors, inter-detector mis-calibration, and intermittent anomalies. Uncalibrated and noisy bands that cover water absorption features were removed, and the remaining 145 bands were included as candidate features. The removed bands are 10-55, 82-97, 102-119, 134-164 and 187-220. The data analyzed in this study, acquired on May 31, 2001 and consist of observations from 14 identified classes representing the land cover types in seasonal swamps, occasional swamps and drier woodlands located in the distal portion of Delta. The selected dataset class description is provided in the Table~\ref{Tab.6} and the ground truth map is presented in Fig.~\ref{Fig.10}.

\begin{table}[!ht]
\begin{minipage}[t]{0.4\linewidth}
\centering
\includegraphics[scale=0.35]{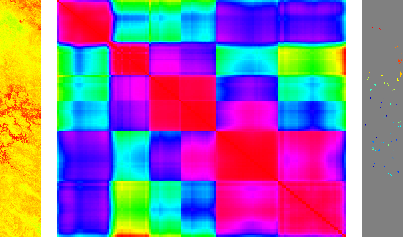}
\captionof{figure}{\textbf{Kennedy Space Centre} Dataset.}
\label{Fig.10}
\end{minipage}
\begin{minipage}[b]{0.6\linewidth}
\centering
\tiny
\captionsetup{justification=centering}
\centering
\caption{Botswana Dataset.}
\begin{tabular}{lcc}
\hline
\textbf{Class Name}    & \textbf{Number of Samples} \\ \hline
Water  & 270\\ \hline
Hippo grass &  101\\ \hline
Floodplain grasses1 &  251\\ \hline
Floodplain grasses2 &  215\\ \hline
Reeds1 &  269\\ \hline
Riparian &  269\\ \hline
Firescar2 &  259\\ \hline
Island interior &  203\\ \hline
Acacia woodlands &  314\\ \hline
Acacia shrublands &  248\\ \hline
Acacia grasslands &  305\\ \hline
Short mopane &  181\\ \hline
Mixed mopane &  268\\ \hline
Exposed soils &  95\\ \hline
\label{Tab.6}
\end{tabular}
\end{minipage}\hfill
\end{table}

\begin{figure}[!ht]
	\centering
	\begin{subfigure}[b]{0.22\textwidth}
		\centering
		\includegraphics[scale=0.4]{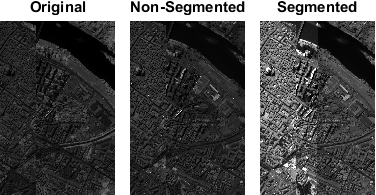}
		\caption[Network2]%
		{{\small Pavia Centre}}
		\label{}
	\end{subfigure}
	\quad
	\begin{subfigure}[b]{0.2\textwidth}  
		\centering 
		\includegraphics[scale=0.27]{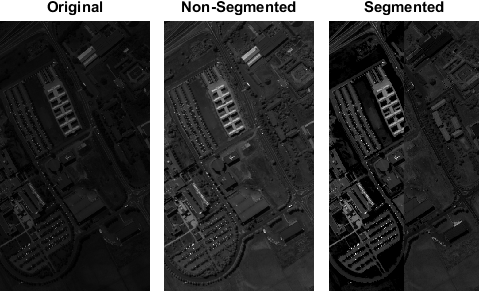}
		\caption[]%
		{{\small Pavia University}}    
		\label{}
	\end{subfigure}
	\vskip\baselineskip
	\begin{subfigure}[b]{0.2\textwidth}   
		\centering 
		\includegraphics[scale=0.35]{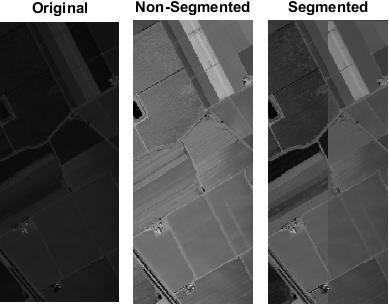}
		\caption[]%
		{{\small Salinas}}    
		\label{}
	\end{subfigure}
	\quad
	\begin{subfigure}[b]{0.2\textwidth}   
		\centering 
		\includegraphics[scale=0.4]{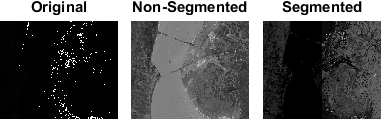}
		\caption[]%
		{{\small Kennedy Space Centre}}    
		\label{}
	\end{subfigure}
	\vskip\baselineskip
	\begin{subfigure}[b]{0.2\textwidth}   
		\centering 
		\includegraphics[scale=0.4]{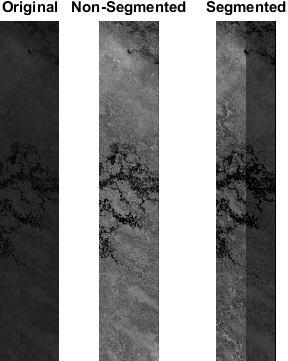}
		\caption[]%
		{{\small Botswana}}    
		\label{}
	\end{subfigure}
	\caption[The input band and the output of \textit{UDAE} on single band in reduced dimension using both segmented and non-segmented \textit{UDAE} method over PC, PU, Salinas, KSC and Botswana Datasets, respectively.]
	{\small The input band and the output of \textit{UDAE} on single band in reduced dimension using both segmented and non-segmented \textit{UDAE} method over PC, PU, Salinas, KSC and Botswana Datasets, respectively.} 
	\label{Fig.11}
\end{figure}

The inner most representation i.e. the last layer of encoding part is visualize in Fig~\ref{Fig.11} as a compression of the input space containing the key contents of hyperspectral bands. Furthermore, Figs~\ref{Fig.11} also shows the ability of \textit{UDAE} method to preserve the key information and original structure of hyperspectral data in lower dimensions for both segmented and non-segmented \textit{UDAE}.

Tables~\ref{Tab.7}-\ref{Tab.9} show the classification accuracy obtained by SVM, kNN, GB, LB, AB and Bag classifiers on PC, PU, Salinas, KSC and Botswana Datasets respectively. From results we conclude that the instance-based classifiers do not lead to better performance as compared to the ensemble learning classifiers because no spatial structure of hyperspectral data have been considered. It also might be the case that the training samples were randomly selected that lead to the class imbalance issue. In our experiments, we randomly chose 30\% of samples per class for training model and the remaining samples were used as a test data.

\begin{table*}[!ht]
	\caption{Classification Performance on Pavia Centre (PC) and Pavia University (PU) Datasets.}
	\centering
	\tiny
	\begin{tabular}[t]{lcc} 
		\multicolumn{3}{c}{\textbf{kappa Coefficient for PC}} \\ \hline
		\textbf{Classifiers} & \textbf{Non-Seg-UDAE} & \textbf{Seg-UDAE} \\ \hline
		SVM & 0.9673 & 0.9735 \\ \hline
		KNN & 0.9680 & 0.9725 \\ \hline
		BAG & 0.9691 & 0.9745 \\ \hline
		AB & 0.9674 & 0.9746 \\ \hline
		LB & 0.9638 & 0.9749 \\ \hline	
		GB & 0.9677 & 0.9744 \\ \hline
	\end{tabular}%
	\quad \
	\begin{tabular}[t]{lcc}
		\multicolumn{3}{c}{\textbf{Overall Classification Accuracy for PC.}} \\ \hline
		\textbf{Classifiers} & \textbf{Non-Seg-UDAE} & \textbf{Seg-UDAE} \\ \hline
		SVM & 0.9769 & 0.9813 \\ \hline
		KNN & 0.9774 & 0.9806 \\ \hline
		BAG & 0.9782 & 0.9819 \\ \hline
		AB  & 0.9769  & 0.9821\\ \hline
		LB  & 0.9745 & 0.9822 \\ \hline	
		GB & 0.9772 & 0.9819\\ \hline
	\end{tabular}
	\quad \
	\begin{tabular}[t]{lcc} 
		\multicolumn{3}{c}{\textbf{kappa Coefficient for PU.}} \\ \hline
		\textbf{Classifiers} & \textbf{Non-Seg-UDAE} & \textbf{Seg-UDAE} \\ \hline
		SVM & 0.7822 & 0.8549 \\ \hline
		KNN & 0.7657 & 0.8474 \\ \hline
		BAG & 0.7704 & 0.8576 \\ \hline
		AB & 0.7722 & 0.8549  \\ \hline
		LB & 0.7772 & 0.8597 \\ \hline	
		GB & 0.7732 & 0.8537 \\ \hline
	\end{tabular}%
	\quad \
	\begin{tabular}[t]{lcc}
		\multicolumn{3}{c}{\textbf{Overall Classification Accuracy for PU.}} \\ \hline
		\textbf{Classifiers} & \textbf{Non-Seg-UDAE} & \textbf{Seg-UDAE} \\ \hline
		SVM & 0.8407 & 0.8920 \\ \hline
		KNN & 0.8289 & 0.8868 \\ \hline
		BAG & 0.8301 & 0.8932 \\ \hline
		AB & 0.8319 & 0.8915 \\ \hline
		LB & 0.8358 & 0.8949 \\ \hline	
		GB & 0.8325 & 0.8903 \\ \hline
	\end{tabular}
	\label{Tab.7}
\end{table*}

\begin{table*}[!ht]
	\caption{Classification Performance on Salinas and Kennedy Space Center (KSC) Datasets.}
	\centering
	\tiny
	\begin{tabular}[t]{lcc} 
		\multicolumn{3}{c}{\textbf{kappa Coefficient for Salinas.}} \\ \hline
		\textbf{Classifiers} & \textbf{Non-Seg-UDAE} & \textbf{Seg-UDAE} \\ \hline
		SVM & 0.9038 & 0.9053 \\ \hline
		KNN & 0.8896 & 0.8937 \\ \hline
		BAG & 0.9118 & 0.9099 \\ \hline
		AB & 0.9079 & 0.9058  \\ \hline
		LB & 0.9103 & 0.9077 \\ \hline	
		GB & 0.9083 & 0.9065 \\ \hline
	\end{tabular}%
	\quad \
	\begin{tabular}[t]{lcc}
		\multicolumn{3}{c}{\textbf{Overall Classification Accuracy for Salinas.}} \\ \hline
		\textbf{Classifiers} & \textbf{Non-Seg-UDAE} & \textbf{Seg-UDAE} \\ \hline
		SVM & 0.9137 & 0.9150 \\ \hline
		KNN & 0.9010 & 0.9047 \\ \hline
		BAG & 0.9208 & 0.9192 \\ \hline
		AB & 0.9174 & 0.9154 \\ \hline
		LB & 0.9195 & 0.9172 \\ \hline	
		GB & 0.9178 & 0.9161 \\ \hline
	\end{tabular}
	\quad \
	\begin{tabular}[t]{lcc} 
		\multicolumn{3}{c}{\textbf{kappa Coefficient for KSC.}} \\ \hline
		\textbf{Classifiers} & \textbf{Non-Seg-UDAE} & \textbf{Seg-UDAE} \\ \hline
		SVM & 0.7255 & 0.7367 \\ \hline
		KNN & 0.7049 & 0.6843 \\ \hline
		BAG & 0.7164 & 0.7630 \\ \hline
		AB & 0.7451 & 0.8056  \\ \hline
		LB & 0.7529 & 0.8099 \\ \hline	
		GB & 0.7542 & 0.8039 \\ \hline
	\end{tabular}%
	\quad \
	\begin{tabular}[t]{lcc}
		\multicolumn{3}{c}{\textbf{Overall Classification Accuracy for KSC..}} \\ \hline
		\textbf{Classifiers} & \textbf{Non-Seg-UDAE} & \textbf{Seg-UDAE} \\ \hline
		SVM & 0.7534 & 0.7636 \\ \hline
		KNN & 0.7353 & 0.7169 \\ \hline
		BAG & 0.7452 & 0.7872 \\ \hline
		AB & 0.7710 & 0.8254 \\ \hline
		LB & 0.7781 & 0.8292 \\ \hline	
		GB & 0.7792 & 0.8239 \\ \hline
	\end{tabular}
	\label{Tab.8}
\end{table*}

\begin{table*}[!ht]
	\caption{Classification Performance on Botswana Dataset.}
	\centering
	\tiny
	\begin{tabular}[t]{lcc} 
		\multicolumn{3}{c}{\textbf{kappa Coefficient}} \\ \hline
		\textbf{Classifiers} & \textbf{Non-Seg-UDAE} & \textbf{Seg-UDAE} \\ \hline
		SVM & 0.8385 & 0.8744 \\ \hline
		KNN & 0.8199 & 0.8629 \\ \hline
		BAG & 0.8509 & 0.8553 \\ \hline
		AB & 0.8451 & 0.8433  \\ \hline
		LB & 0.8408 & 0.8361 \\ \hline	
		GB & 0.8484 & 0.8380 \\ \hline
	\end{tabular}%
	\quad \
	\begin{tabular}[t]{lcc}
		\multicolumn{3}{c}{\textbf{Overall Classification Accuracy.}} \\ \hline
		\textbf{Classifiers} & \textbf{Non-Seg-UDAE} & \textbf{Seg-UDAE} \\ \hline
		SVM & 0.8509 & 0.8839 \\ \hline
		KNN & 0.8337 & 0.8734 \\ \hline
		BAG & 0.8624 & 0.8663 \\ \hline
		AB & 0.8571 & 0.8553 \\ \hline
		LB & 0.8531 & 0.8487 \\ \hline	
		GB & 0.8602 & 0.8505 \\ \hline
		\label{Tab.9}
	\end{tabular}
\end{table*}

\begin{figure*}[!ht]
	\centering
	\begin{subfigure}[b]{0.3\textwidth}
		\centering
		\includegraphics[scale=0.3]{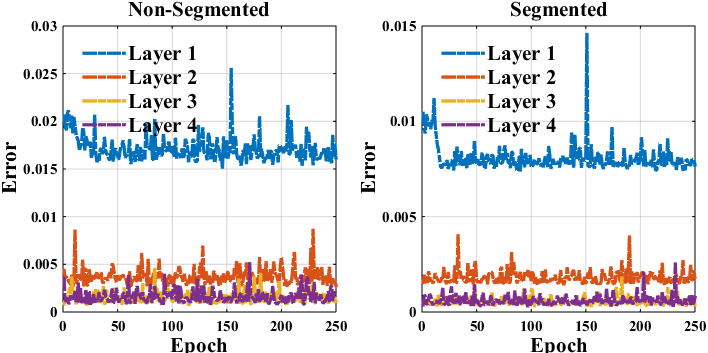}
		\caption[Network2]%
		{{\small Pavia Centre}}
		\label{}
	\end{subfigure}
	\quad
	\begin{subfigure}[b]{0.3\textwidth}  
		\centering 
		\includegraphics[scale=0.3]{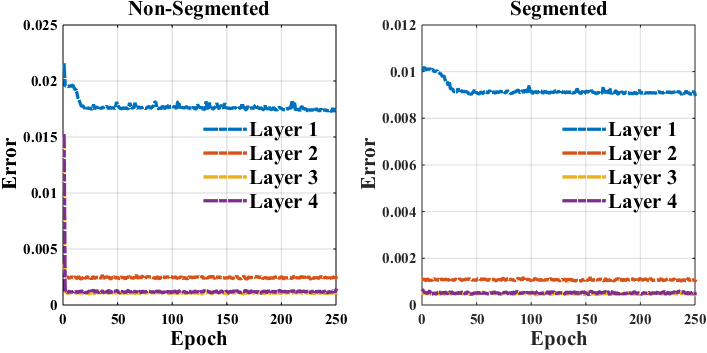}
		\caption[]%
		{{\small Pavia University}}    
		\label{}
	\end{subfigure}
	\begin{subfigure}[b]{0.33\textwidth}   
		\centering 
		\includegraphics[scale=0.3]{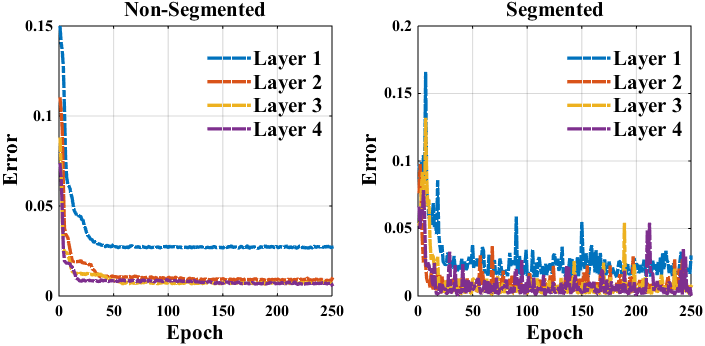}
		\caption[]%
		{{\small Salinas}}    
		\label{}
	\end{subfigure}
	\vskip\baselineskip
	\begin{subfigure}[b]{0.45\textwidth}   
		\centering 
		\includegraphics[scale=0.3]{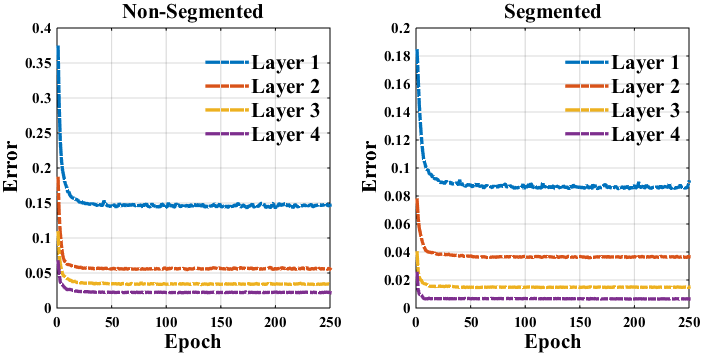}
		\caption[]%
		{{\small Kennedy Space Centre}}    
		\label{}
	\end{subfigure}
	\quad
	\begin{subfigure}[b]{0.45\textwidth}   
		\centering 
		\includegraphics[scale=0.3]{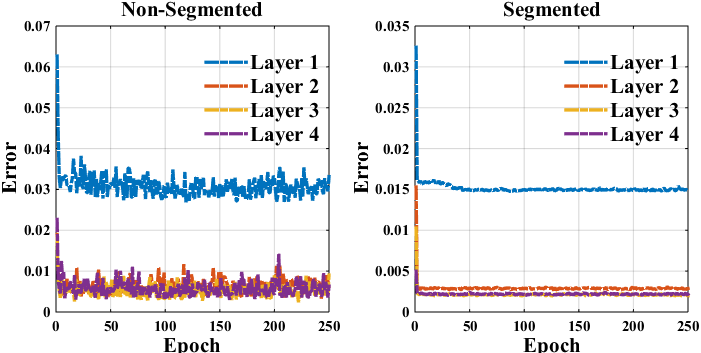}
		\caption[]%
		{{\small Botswana}}    
		\label{}
	\end{subfigure}
	\caption[Reconstruction Error on Each Layer corresponding to the number of iterations on each layer for both Segmented and non-segmented \textit{UDAE} method over PC, PU, Salinas, KSC and Botswana Datasets, respectively.]
	{\small Reconstruction Error on Each Layer corresponding to the number of iterations on each layer for both Segmented and non-segmented \textit{UDAE} method over PC, PU, Salinas, KSC and Botswana Datasets, respectively.} 
	\label{Fig.12}
\end{figure*}

The sensitivity analysis of reconstruction of actual data from the DAE network based on the provided settings explained in section III-B (main paper). Fig~\ref{Fig.12} shows the results of reconstruction error on each layer with respect to the number of epochs on each layer for both Segmented and non-segmented \textit{UDAE}.

\begin{figure*}[!ht]
	\centering
	\begin{subfigure}[b]{0.45\textwidth}
		\centering
		\includegraphics[scale=0.48]{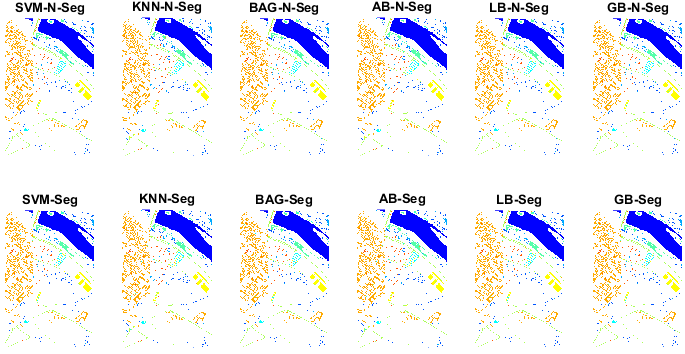}
		\caption[Network2]%
		{{\small Pavia Centre}}
		\label{}
	\end{subfigure}
	\quad
	\begin{subfigure}[b]{0.45\textwidth}  
		\centering 
		\includegraphics[scale=0.4]{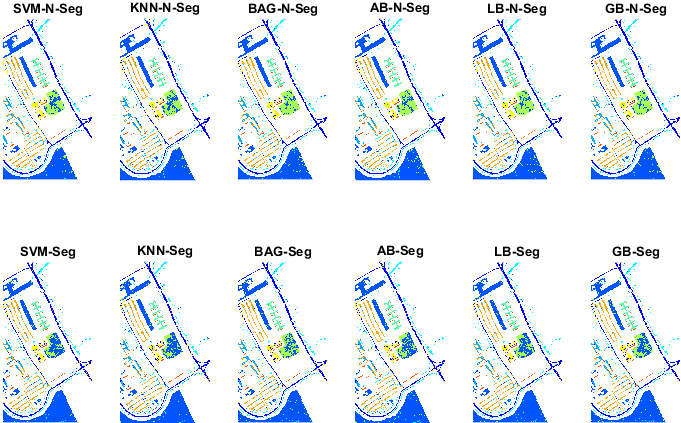}
		\caption[]%
		{{\small Pavia University}}    
		\label{}
	\end{subfigure}
	\vskip\baselineskip
	\begin{subfigure}[b]{0.45\textwidth}   
		\centering 
		\includegraphics[scale=0.45]{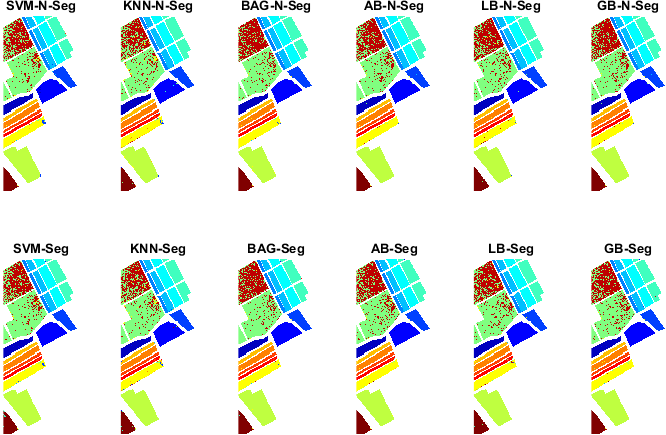}
		\caption[]%
		{{\small Salinas}}    
		\label{}
	\end{subfigure}
	\quad
	\begin{subfigure}[b]{0.45\textwidth}   
		\centering 
		\includegraphics[scale=0.45]{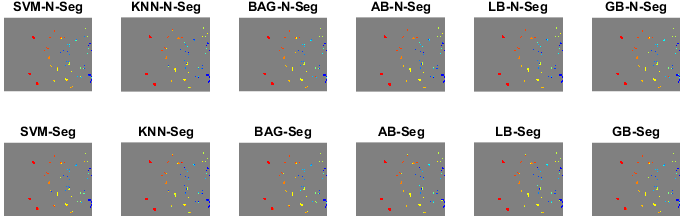}
		\caption[]%
		{{\small Kennedy Space Centre}}    
		\label{}
	\end{subfigure}
	\vskip\baselineskip
	\begin{subfigure}[b]{0.45\textwidth}   
		\centering 
		\includegraphics[scale=0.5]{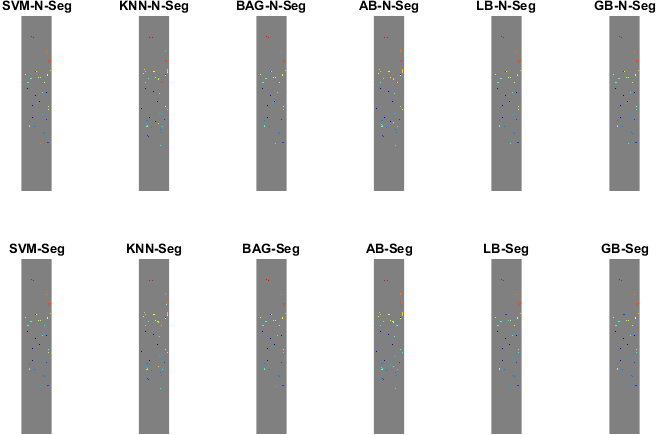}
		\caption[]%
		{{\small Botswana}}    
		\label{}
	\end{subfigure}
	\caption[Reconstruction Error on Each Layer corresponding to the number of iterations on each layer for both Segmented and non-segmented \textit{UDAE} method over PC, PU, Salinas, KSC and Botswana Datasets, respectively.]
	{\small Reconstruction Error on Each Layer corresponding to the number of iterations on each layer for both Segmented and non-segmented \textit{UDAE} method over PC, PU, Salinas, KSC and Botswana Datasets, respectively.} 
	\label{Fig.13}
\end{figure*}

Fig~\ref{Fig.13} shows the visual classification results of the six classifiers e.g. SVM, kNN, GB, LB, AB and Bag respectively. From results, we observe that the segmented \textit{UDAE} method slightly perform better in terms of defining the classes then non-segmented \textit{UDAE}. Segmented \textit{UDAE} breaks the similar clusters into several parts which helps to obtain better classification and generalization performance in computationally efficient fashion then the non-segmented \textit{UDAE}. The obtained results demonstrate the acceptable generalization capability for the \textit{UDAE} method for all these three types of classifiers.

\begin{figure*}[!ht]
	\centering
	\begin{subfigure}[b]{0.45\textwidth}
		\centering
		\includegraphics[scale=0.3]{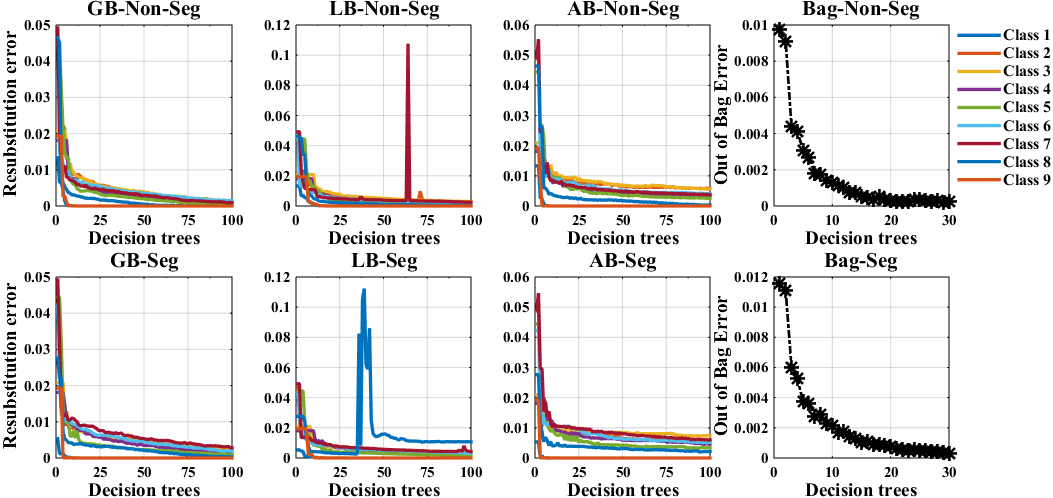}
		\caption[Network2]%
		{{\small Pavia Centre}}
		\label{}
	\end{subfigure}
	\quad
	\begin{subfigure}[b]{0.45\textwidth}  
		\centering 
		\includegraphics[scale=0.3]{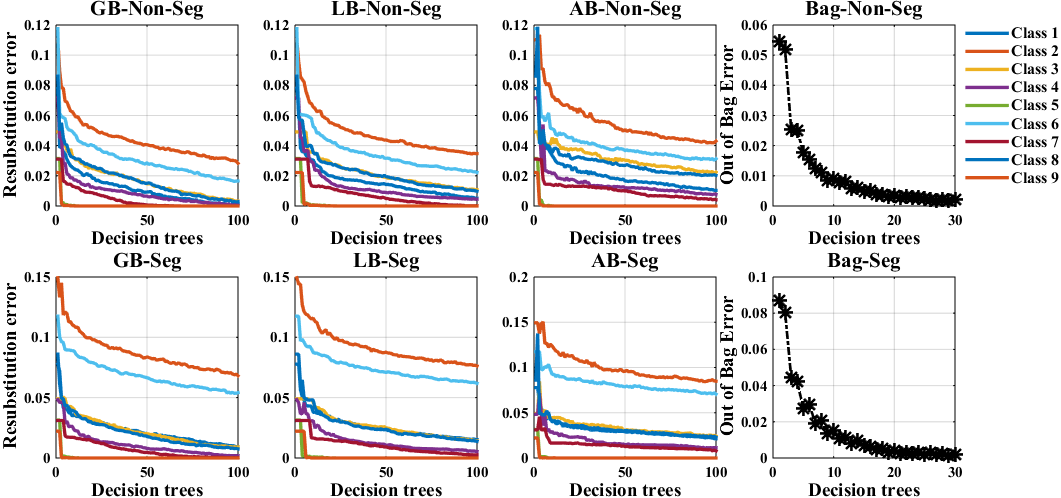}
		\caption[]%
		{{\small Pavia University}}    
		\label{}
	\end{subfigure}
	\vskip\baselineskip
	\begin{subfigure}[b]{0.45\textwidth}   
		\centering 
		\includegraphics[scale=0.3]{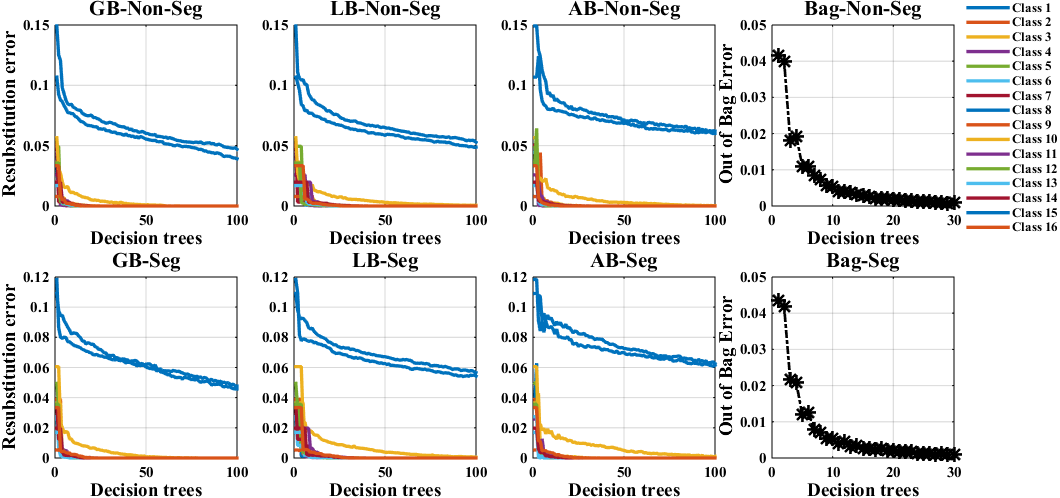}
		\caption[]%
		{{\small Salinas}}    
		\label{}
	\end{subfigure}
	\quad
	\begin{subfigure}[b]{0.45\textwidth}   
		\centering 
		\includegraphics[scale=0.3]{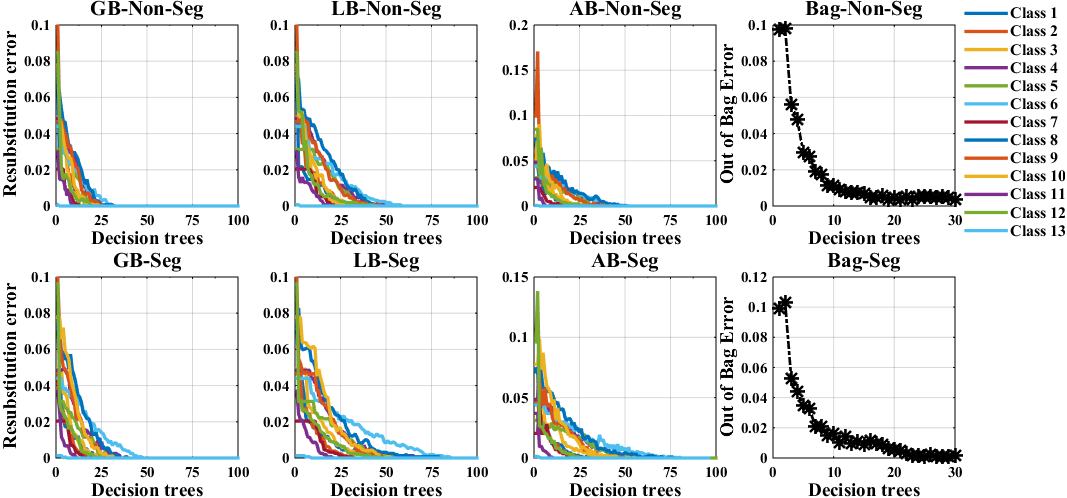}
		\caption[]%
		{{\small Kennedy Space Centre}}    
		\label{}
	\end{subfigure}
	\vskip\baselineskip
	\begin{subfigure}[b]{0.45\textwidth}   
		\centering 
		\includegraphics[scale=0.3]{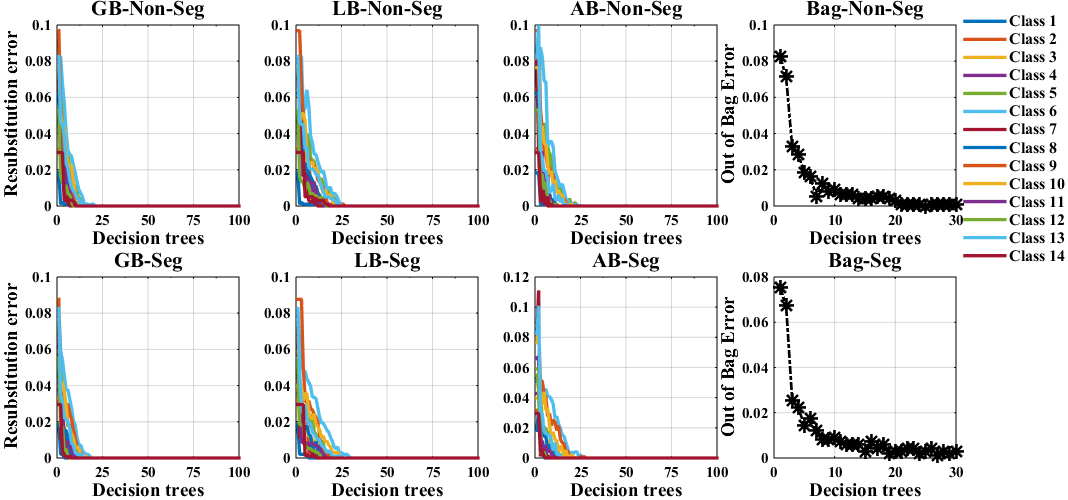}
		\caption[]%
		{{\small Botswana}}    
		\label{}
	\end{subfigure}
	\caption[Reconstruction Error on Each Layer corresponding to the number of iterations on each layer for both Segmented and non-segmented \textit{UDAE} method over PC, PU, Salinas, KSC and Botswana Datasets, respectively.]
	{\small Reconstruction Error on Each Layer corresponding to the number of iterations on each layer for both Segmented and non-segmented \textit{UDAE} method over PC, PU, Salinas, KSC and Botswana Datasets, respectively.} 
	\label{Fig.14}
\end{figure*}

In this experiment, we analyze the behavior of different ensemble learning classifiers based on resubstitution and out-of-bag error against different number of decision trees. Fig~\ref{Fig.14} shows the results of resubstitution and out-of-bag error for GB, LB, AB and Bag classifiers respectively.

\end{document}